\documentclass[12pt]{article}
\renewenvironment{abstract}{\centerline{\bf
Abstract}\vspace{0.5ex}\begin{quote}\small}{\par\end{quote}\vskip
1ex}
\newenvironment{keywords}{\centerline{\bf
Keywords}\vspace{0.5ex}\begin{quote}\small}{\par\end{quote}\vskip
1ex}
\newtheorem{theorem}{Theorem}

\def\gtapprox{\buildrel{\lower.7ex\hbox{$>$}}\over
                       {\lower.7ex\hbox{$\sim$}}}
\def\nq{\hspace{-1em}}

\def\ignore#1{}

\def\qed{\sqcap\!\!\!\!\sqcup}
\def\odt{{\textstyle{1\over 2}}}
\def\odf{{\textstyle{1\over 4}}}
\def\odA{{\textstyle{1\over A}}}
\def\hbar{h\!\!\!\!^{-}\,}

\def\beq{\begin{equation}}
\def\eeq{\end{equation}}
\def\beqn{\begin{displaymath}}
\def\eeqn{\end{displaymath}}
\def\bqa{\begin{equation}\begin{array}{c}}
\def\eqa{\end{array}\end{equation}}
\def\bqan{\begin{displaymath}\begin{array}{c}}
\def\eqan{\end{array}\end{displaymath}}
\def\pb{\underline}                       
\def\pb#1{\underline{#1}}                 
\def\maxarg{\mathop{\rm maxarg}}          
\def\minarg{\mathop{\rm minarg}}          

\begin{document}


\begin{titlepage}


\begin{center}\vspace*{-1cm}
  {\small Technical Report IDSIA-07-01, 26. February 2001} \\[1cm]
  {\LARGE\sc\hrule height1pt \vskip 4mm Convergence and Error Bounds \\
             for Universal Prediction \\
  of Nonbinary Sequences\vskip 3mm \hrule height1pt} \vspace{1.5cm}
  {\bf Marcus Hutter}                    \\[1cm]
  {\rm IDSIA, Galleria 2, CH-6928 Manno-Lugano, Switzerland}  \\
  {\rm\footnotesize marcus@idsia.ch
   \footnote{This work was supported by SNF grant 2000-61847.00 to J\"urgen Schmidhuber.}
   \qquad http://www.idsia.ch/$^{_{_\sim}}\!$marcus} \\[1cm]
\end{center}



\begin{abstract}
Solomonoff's uncomputable universal prediction scheme $\xi$ allows
to predict the next symbol $x_k$ of a sequence $x_1...x_{k-1}$ for
any Turing computable, but otherwise unknown, probabilistic
environment $\mu$. This scheme will be generalized to arbitrary
environmental classes, which, among others, allows the
construction of computable universal prediction schemes $\xi$.
Convergence of $\xi$ to $\mu$ in a conditional mean squared sense
and with $\mu$ probability $1$ is proven. It is shown that the
average number of prediction errors made by the universal $\xi$
scheme rapidly converges to those made by the best possible
informed $\mu$ scheme. The schemes, theorems and proofs are given
for general finite alphabet, which results in additional
complications as compared to the binary case. Several extensions
of the presented theory and results are outlined. They include
general loss functions and bounds, games of chance, infinite
alphabet, partial and delayed prediction, classification, and more
active systems.
\end{abstract}

\end{titlepage}

\section{Introduction}\label{secInt}

The Bayesian framework is ideally suited for studying induction
problems. The probability of observing $x_k$ at time $k$, given
past observations $x_1...x_{k-1}$, can be computed with Bayes'
rule if the generating probability distribution $\mu$, from which
sequences $x_1x_2x_3...$ are drawn, is known. The problem,
however, is that in many cases one does not even have a reasonable
estimate of the true generating distribution. What is the true
probability of weather sequences or stock charts?
In order to overcome this problem we define a universal
distribution $\xi$ as a weighted sum of distributions
$\mu_i\!\in\!M$, where $M$ is any finite or countable set of
distributions including $\mu$. This is a generalization of
Solomonoff induction, in which $M$ is the set of all enumerable
semi-measures \cite{Solomonoff:64,Solomonoff:78}.
We show that using the universal $\xi$ as a prior is nearly as
good as using the unknown generating distribution $\mu$. In a sense,
this solves the problem, that the generating distribution $\mu$ is not
known, in a universal way. All results are obtained for general
finite alphabet.
Convergence of $\xi$ to $\mu$ in a conditional mean squared sense
and with $\mu$ probability $1$ is proven. The number of
errors $E_{\Theta_\xi}$ made by the universal prediction
scheme $\Theta_\xi$ based on $\xi$ minus the number of errors
$E_{\Theta_\mu}$ of the optimal informed prediction scheme
$\Theta_\mu$ based on $\mu$ is proven to be bounded by
$O(\sqrt{E_{\Theta_\mu}})$.

Extensions to arbitrary loss functions, games of chance, infinite
alphabet, partial and delayed prediction, classification, and
more active systems are discussed (Section \ref{secOut}).
The main new results are a generalization of the universal
probability $\xi$ \cite{Solomonoff:64} to arbitrary probability
classes and weights (Section \ref{secSetup}), a generalization of
the convergence \cite{Solomonoff:78} $\xi\to\mu$ (Section
\ref{secConv}) and the error bounds \cite{Hutter:99} to arbitrary
alphabet (Section \ref{secErr}). The non-binary setting causes
substantial additional complications. Non-binary prediction cannot
be (easily) reduced to the binary case. One may have in mind a
binary coding of the symbols $x_k$ in the sequence
$x_1x_2...$. But this makes it necessary to predict a block of
bits $x_k$, before receiving the true block of bits $x_k$, which
differs from the bit-by-bit prediction considered in
\cite{Solomonoff:78,Li:97,Hutter:99}.

For an excellent introduction to Kolmogorov complexity and
Solomonoff induction one should consult the book of Li and
Vit\'anyi \cite{Li:97} or the article \cite{Li:92b} for a short
course. Historical surveys of inductive reasoning and inference
can be found in \cite{Angluin:83,Solomonoff:97}.

\section{Setup}\label{secSetup}

\subsection{Strings and Probability Distributions}
We denote strings over a finite alphabet $\cal A$ by
$x_1x_2...x_n$ with $x_k\!\in\!\cal A$. We further use the
abbreviations $x_{n:m}:=x_nx_{n+1}...x_{m-1}x_m$ and
$x_{<n}:=x_1... x_{n-1}$. We use Greek letters for probability
distributions and underline their arguments to indicate that they
are probability arguments. Let $\rho(\pb{x_1...x_n})$ be the
probability that an (infinite) sequence starts with $x_1...x_n$:
\beq\label{prop}
  \sum_{x_{1:n}\in{\cal A}^n}\rho(\pb x_{1:n})=1,\quad
  \sum_{x_n\in\cal A}\rho(\pb x_{1:n}) =
  \rho(\pb x_{<n})
  ,\quad
  \rho(\epsilon)=1.
\eeq We also need conditional probabilities derived from Bayes'
rule. We prefer a notation which preserves the order of the words,
in contrast to the standard notation $\rho(\cdot|\cdot)$ which
flips it. We extend the definition of $\rho$ to the conditional
case with the following convention for its arguments: An
underlined argument $\pb x_k$ is a probability variable and other
non-underlined arguments $x_k$ represent conditions. With this
convention, Bayes' rule has the following look:
\beq\label{bayes}
  \rho(x_{<n}\pb x_n) =
  \rho(\pb x_{1:n})/\rho(\pb x_{<n}) \quad,\quad
  \rho({\pb{x_1...x_n}}) =
  \rho(\pb x_1)\!\cdot\!
  \rho(x_1\pb x_2)
  \!\cdot\!...\!\cdot\!
  \rho(x_1...x_{n-1}\pb x_n).
\eeq
The first equation states that the probability that a string
$x_1...x_{n-1}$ is followed by $x_n$ is equal to the probability
that a string starts with $x_1...x_n$ divided by the probability
that a string starts with $x_1...x_{n-1}$. The second equation is
the first, applied $n$ times.

\subsection{Universal Prior Probability Distribution}
Most inductive inference problem can be brought into the
following form: Given a string $x_{<k}$, take a guess at its
continuation $x_k$. We will assume that the strings which have to
be continued are drawn from a probability\footnote{This includes
deterministic environments, in which case the probability
distribution $\mu$ is $1$ for some sequence $x_{1:\infty}$ and $0$
for all others. We call probability distributions of this kind
{\it deterministic}.} distribution $\mu$. The maximal prior
information a prediction algorithm can possess is the exact
knowledge of $\mu$, but in many cases
the generating distribution is not known. Instead, the prediction is
based on a guess $\rho$ of $\mu$. We expect that a predictor based
on $\rho$ performs well, if $\rho$ is close to $\mu$ or converges,
in a sense, to $\mu$. Let $M\!:=\!\{\mu_1,\mu_2,...\}$ be a finite
or countable set of candidate probability distributions on
strings. We define a weighted average on $M$
\beq\label{xidef}
  \xi(\pb x_{1:n}) \;:=\;
  \sum_{\mu_i\in M}w_{\mu_i}\!\cdot\!\mu_i(\pb x_{1:n}),\quad
  \sum_{\mu_i\in M}w_{\mu_i}=1,\quad w_{\mu_i}>0.
\eeq
It is easy to see that $\xi$ is a probability distribution as
the weights $w_{\mu_i}$ are positive and normalized to 1 and the
$\mu_i\!\in\!M$ are probabilities. For finite $M$ a possible
choice for the $w$ is to give all $\mu_i$ equal weight
($w_{\mu_i}={1\over|M|}$). We call $\xi$ universal relative to
$M$, as it multiplicatively dominates all distributions in $M$
\beq\label{unixi}
  \xi(\pb x_{1:n}) \;\geq\;
  w_{\mu_i}\!\cdot\!\mu_i(\pb x_{1:n}) \quad\mbox{for all}\quad
  \mu_i\in M.
\eeq
In the following, we assume that $M$ is known and
contains\footnote{Actually all theorems remain valid for $\mu$ being a finite linear
combination of $\mu_i\in L\subseteq M$ and
$w_\mu:=\min_{\mu_i\in L}w_{\mu_i}$ \cite{Hutter:01op}.}
the true generating distribution, i.e. $\mu\!\in\!M$. We will see that this
is not a serious constraint as we can always chose $M$ to be
sufficiently large. In the next section we show the important
property of $\xi$ converging to the generating distribution $\mu$ in a
sense and, hence, might being a useful substitute for the true generating,
but in general, unknown distribution $\mu$.

\subsection{Probability Classes}
We get a rather wide class $M$ if we include {\it all} computable
probability distributions in $M$. In this case, the assumption
$\mu\!\in\!M$ is very weak, as it only assumes that the strings
are drawn from {\it any computable} distribution; and all valid
physical theories (and, hence, all environments) {\it are}
computable (in a probabilistic sense).

We will see that it is favorable to assign high weights
$w_{\mu_i}$ to the $\mu_i$. Simplicity should be favored over
complexity, according to Occam's razor. In our context this means
that a high weight should be assigned to simple $\mu_i$. The
prefix Kolmogorov complexity $K(\mu_i)$ is a universal complexity
measure \cite{Kolmogorov:65,Zvonkin:70,Li:97}. It is defined as
the length of the shortest self-delimiting program (on a universal
Turing machine) computing $\mu_i(x_{1:n})$ given $x_{1:n}$. If we
define \beqn
  w_{\mu_i} \;:=\; {1\over\Omega}2^{-K(\mu_i)} \quad,\quad
  \Omega \;:=\; \sum_{\mu_i\in M}2^{-K(\mu_i)}
\eeqn then, distributions which can be calculated by short
programs, have high weights. Besides ensuring correct
normalization, $\Omega$ (sometimes called the number of wisdom)
has interesting properties in itself
\cite{Calude:98,Chaitin:91}. If we enlarge $M$ to
include all enumerable semi-measures, we attain Solomonoff's
universal probability, apart from normalization, which has to be
treated differently in this case
\cite{Solomonoff:64,Solomonoff:78}. Recently, $M$ has been further
enlarged to include all cumulatively enumerable semi-measures
\cite{Schmidhuber:01}. In all cases, $\xi$ is not finitely
computable, but can still be approximated to arbitrary but not
pre-specifiable precision. If we consider {\it all} approximable
(i.e.\ asymptotically computable) distributions, then the
universal distribution $\xi$, although still well defined, is not
even approximable \cite{Schmidhuber:01}. An interesting and
quickly approximable distribution is the Speed prior $S$ defined
in \cite{Schmidhuber:01}. It is related to Levin complexity and
Levin search \cite{Levin:73,Levin:84}, but it is unclear for now
which distributions are dominated by $S$. If one considers only
finite-state automata instead of general Turing machines, one can
attain a quickly computable, universal finite-state prediction
scheme related to that of Feder et al. \cite{Feder:92}, which
itself is related to the famous Lempel-Ziv data compression
algorithm. If one has extra knowledge on the source generating the
sequence, one might further reduce $M$ and increase $w$. A
detailed analysis of these and other specific classes $M$ will be
given elsewhere. Note that $\xi\!\in\!M$ in the enumerable and
cumulatively enumerable case, but $\xi\!\not\in\!M$ in the
computable, approximable and finite-state case. If $\xi$ is itself
in $M$, it is called a universal element of $M$ \cite{Li:97}. As
we do not need this property here, $M$ may be {\it any} finite or
countable set of distributions. In the following we consider
generic $M$ and $w$.

\section{Convergence}\label{secConv}

\subsection{Upper Bound for the Relative Entropy}
Let us define the relative entropy (also called Kullback Leibler
divergence \cite{Kullback:59}) between $\mu$ and $\xi$:
\beq\label{hn}
  h_k(x_{<k}) \;:=\; \sum_{x_k\in\cal A}\mu(x_{<k}\pb x_k)
  \ln{\mu(x_{<k}\pb x_k) \over \xi(x_{<k}\pb x_k)}.
\eeq
$H_n$ is then defined as the sum-expectation, for which the
following upper bound can be shown
\beq\label{entropy}
  H_n \;:=\; \sum_{k=1}^n \nq\nq\;\sum_{\qquad x_{<k}\in{\cal
  A}^{k-1}}\nq\nq\;
  \mu(\pb x_{<k})\!\cdot\!h_k(x_{<k}) \;=\;
  \sum_{k=1}^n\nq\;\sum_{\quad x_{1:k}\in{\cal A}^k}\nq\;\mu(\pb x_{1:k})
  \ln{\mu(x_{<k}\pb x_k)\over\xi(x_{<k}\pb x_k)} \;=
\eeq
\beqn
  =\;
  \sum_{x_{1:n}}\mu(\pb x_{1:n})
  \ln \prod_{k=1}^n{\mu(x_{<k}\pb x_k)\over\xi(x_{<k}\pb x_k)}
  \;=\;
  \sum_{x_{1:n}} \mu(\pb x_{1:n})
  \ln{\mu(\pb x_{1:n}) \over \xi(\pb x_{1:n})} \;\;\leq\;\;
  \ln{1\over w_\mu} \;=:\; d_\mu
\eeqn
In the first line we have inserted (\ref{hn}) and used Bayes' rule
$\mu(\pb x_{<k})\!\cdot\!\mu(x_{<k}\pb x_k)\!=\!\mu(\pb x_{1:k})$.
Due to (\ref{prop}), we can further replace $\sum_{x_{1:k}}\mu(\pb
x_{1:k})$ by $\sum_{x_{1:n}}\mu(\pb x_{1:n})$ as the argument of
the logarithm is independent of $x_{k+1:n}$. The $k$ sum can now
be exchanged with the $x_{1:n}$ sum and transforms to a product
inside the logarithm. In the last equality we have used the second
form of Bayes' rule (\ref{bayes}) for $\mu$ and $\xi$. Using
universality (\ref{unixi}) of $\xi$, i.e.\  $\ln \mu(\pb
x_{1:n})/\xi(\pb x_{1:n})\!\leq\!\ln{1\over w_\mu}$ for
$\mu\!\in\!M$ yields the final inequality in (\ref{entropy}). The
proof given here is simplified version of those given in
\cite{Solomonoff:78} and \cite{Li:97}.

\subsection{Lower Bound for the Relative Entropy}
We need the following inequality to lower bound $H_n$
\beq\label{entroineqn}\label{entro2}
  \sum_{i=1}^N (y_i\!-\!z_i)^2 \;\leq\;
  \sum_{i=1}^N y_i\ln{y_i\over z_i} \quad\mbox{for}\quad
  y_i\geq 0, \quad z_i\geq 0, \quad
  \sum_{i=1}^N y_i=1=\sum_{i=1}^N z_i.
\eeq
The proof of the case $N\!=\!2$
\beq\label{entroineq2}
  2(y\!-\!z)^2 \;\leq\; y\ln{y\over z}+(1\!-\!y)\ln{1\!-\!y\over 1\!-\!z},\quad
  0<y<1,\quad 0<z<1
\eeq
will not be repeated here, as it is elementary and well
known \cite{Li:97}. The proof of
(\ref{entro2}) is one point where the generalization
from binary to arbitrary alphabet is not
trivial.$\!$\footnote{We will not explicate every subtlety and only
sketch the proofs. Subtleties regarding $y,z=0/1$ have been
checked but will be passed over. $0\ln{0\over z_i}\!:=\!0$ even
for $z_i=0$. Positive means $\geq 0$.}
We will reduce the general case $N\!>\!2$
to the case $N\!=\!2$. We do this by a partition
$\{1,...,N\}=G^+\cup G^-$, $G^+\cap G^-=\{\}$, and define
$\displaystyle y^\pm\!:=\!\sum_{i\in G^\pm}y_i$ and $\displaystyle
z^\pm\!:=\!\sum_{i\in G^\pm}z_i$. It is well known that the
relative entropy is positive, i.e.
\beq\label{entropos}
  \sum_{i\in G^\pm}p_i\ln{p_i\over q_i}\;\geq\; 0 \quad\mbox{for}\quad
  p_i\geq 0, \quad q_i\geq 0, \quad
  \sum_{i\in G^\pm} p_i=1=\sum_{i\in G^\pm} q_i.
\eeq
Note that there are 4 probability distributions ($p_i$ and $q_i$
for $i\!\in\!G^+$ and $i\!\in\!G^-$). For $i\!\in\!G^\pm$,
$p_i:=y_i/y^\pm$ and $q_i:=z_i/z^\pm$ satisfy the conditions on
$p$ and $q$. Inserting this into (\ref{entropos}) and rearranging
the terms we get
$\sum_{i\in G^\pm}y_i\ln{y_i\over z_i}\!\geq\!y^\pm\ln{y^\pm\over
z^\pm}.$
If we sum this over $\pm$ and define $y\equiv y^+=1\!-\!y^-$ and
$z\equiv z^+=1\!-\!z^-$ we get
\beq\label{eirb}
  \sum_{i=1}^N y_i\ln{y_i\over z_i}\;\geq\;
  \sum_\pm y^\pm\ln{y^\pm\over z^\pm} \;=\;
  y\ln{y\over z}+(1\!-\!y)\ln{1\!-\!y\over 1\!-\!z}.
\eeq
For the special choice $G^\pm\!:=\!\{i:y_i{>\atop\leq}z_i\}$,
we can upper bound the quadratic term by
\beqn
  \sum_{i\in G^\pm}(y_i\!-\!z_i)^2 \;\leq\;
  \Big(\sum_{i\in G^\pm}|y_i\!-\!z_i|\Big)^2 \;=\;
  \Big(\sum_{i\in G^\pm}y_i\!-\!z_i\Big)^2 \;=\;
  (y^\pm\!-\!z^\pm)^2.
\eeqn
The first equality is true, since all $y_i\!\!-\!\!z_i$ are
positive/negative for $i\!\in\!G^\pm$ due to the special choice of
$G^\pm$. Summation over $\pm$ gives
\beq\label{sqineq}
  \sum_{i=1}^N (y_i\!-\!z_i)^2 \;\leq\;
  \sum_\pm (y^\pm\!-\!z^\pm)^2 \;=\;
  2(y\!-\!z)^2
\eeq
Chaining the inequalities (\ref{sqineq}), (\ref{entroineq2})
and (\ref{eirb}) proves (\ref{entroineqn}).
If we identify
\beq\label{xydef}
  {\cal A}=\{1,...,N\},\quad
  N=|{\cal A}|, \quad
  i=x_k, \quad
  y_i=\mu(x_{<k}\pb x_k), \quad
  z_i=\xi(x_{<k}\pb x_k)
\eeq multiply both sides of (\ref{entro2}) with $\mu(\pb x_{<k})$
and take the sum over $x_{<k}$ and $k$ we get
\beq\label{eukdistxi}
  \sum_{k=1}^n\sum_{x_{1:k}}\mu(\pb x_{<k})
  \Big(\mu(x_{<k}\pb x_k)-\xi(x_{<k}\pb x_k)\Big)^2 \;\leq\;
  \sum_{k=1}^n\sum_{x_{1:k}}\mu(\pb x_{1:k})
  \ln{\mu(x_{<k}\pb x_k)\over\xi(x_{<k}\pb x_k)}.
\eeq

\subsection{Convergence of $\xi$ to $\mu$}
The upper (\ref{entropy}) and lower (\ref{eukdistxi}) bounds on
$H_n$ allow us to prove the convergence of $\xi$ to $\mu$ in a
conditional mean squared sense and with $\mu$ probability 1.

\addcontentsline{toc}{paragraph}{Theorem \ref{thConv}
(Convergence)}
\begin{theorem}[Convergence]\label{thConv}
Let there be sequences $x_1x_2...$ over a finite alphabet $\cal A$
drawn with probability $\mu(\pb x_{1:n})$ for the first $n$
symbols. The universal conditional probability $\xi(x_{<k}\pb
x_k)$
of the next symbol $x_k$ given $x_{<k}$ 
is related to the generating conditional probability $\mu(x_{<k}\pb
x_k)$ in the following way: $$
\begin{array}{rl}
   i) & \displaystyle
        \sum_{k=1}^n\sum_{x_{1:k}}\mu(\pb x_{<k})
        \Big(\mu(x_{<k}\pb x_k)-\xi(x_{<k}\pb x_k)\Big)^2 \;\leq\;
        H_n \;\leq\; d_\mu \;=\; \ln{1\over w_\mu} \;<\; \infty \\[3ex]
  ii) & \xi(x_{<k}\pb x_k) \to \mu(x_{<k}\pb x_k)
        \quad\mbox{for $k\to\infty$ with $\mu$ probability 1}\quad
\end{array}
$$ where $H_n$ is the relative entropy (\ref{entropy}), and
$w_\mu$ is the weight (\ref{xidef}) of $\mu$ in $\xi$.
\end{theorem}

{\sl$(i)$} follows from (\ref{entropy}) and (\ref{eukdistxi}). For
$n\!\to\!\infty$ the l.h.s.\ of {\sl$(i)$} is an infinite $k$-sum
over positive arguments, which is bounded by the finite constant
$d_\mu$ on the r.h.s. Hence the arguments must converge to zero
for $k\!\to\!\infty$. Since the arguments are $\mu$ expectations
of the squared difference of $\xi$ and $\mu$, this means that
$\xi(x_{<k}\pb x_k)$ converges to $\mu(x_{<k}\pb x_k)$ with $\mu$
probability 1 or, more stringent, in a mean square sense. This
proves {\sl$(ii)$}. The reason for the astonishing property of a
single (universal) function $\xi$ to converge to {\it any}
$\mu_i\!\in\!M$ lies in the fact that the sets of $\mu$-random
sequences differ for different $\mu$. Since the conditional
probabilities are the basis of all prediction algorithms
considered in this work, we expect a good prediction performance
if we use $\xi$ as a guess of $\mu$. Performance measures are
defined in the following sections.

\section{Error Bounds}\label{secErr}

We now consider the following measure for the quality of a
prediction: making a wrong prediction counts as one
error, making a correct prediction counts as no error.

\subsection{Total Expected Numbers of Errors}
Let $\Theta_\mu$ be the optimal prediction scheme when the strings
are drawn from the probability distribution $\mu$, i.e. the
probability of $x_k$ given $x_{<k}$ is $\mu(x_{<k}\pb x_k)$, and
$\mu$ is known. $\Theta_\mu$ predicts (by definition)
$x_k^{\Theta_\mu}$ when observing $x_{<k}$. The prediction is
erroneous if the true k$^{th}$ symbol is not $x_k^{\Theta_\mu}$.
The probability of this event is $1-\mu(x_{<k}\pb
x_k^{\Theta_\mu})$. It is minimized if $x_k^{\Theta_\mu}$
maximizes $\mu(x_{<k}\pb x_k^{\Theta_\mu})$. More generally, let
$\Theta_\rho$ be a prediction scheme predicting
$x_k^{\Theta_\rho}\!:=\!\maxarg_{x_k}\rho(x_{<k}\pb x_k)$ for some
distribution $\rho$. Every deterministic predictor can be
interpreted as maximizing some distribution. The $\mu$ probability
of making a wrong prediction for the $k^{th}$ symbol and the total
$\mu$-expected number of errors in the first $n$ predictions of
predictor $\Theta_\rho$ are
\beq\label{rhoerr}
  e_{k\Theta_\rho}(x_{<k}) \;:=\; 1-\mu(x_{<k}\pb x_k^{\Theta_\rho})
  \quad,\quad
  E_{n\Theta_\rho} \;:=\; \sum_{k=1}^n \nq\;\sum_{\quad x_1...x_{k-1}}\nq
  \mu(\pb x_{<k})\!\cdot\!e_{k\Theta_\rho}(x_{<k}).
\eeq If $\mu$ is known, $\Theta_\mu$ is obviously the best
prediction scheme in the sense of making the least number of
expected errors \beq\label{Emuopt}
  E_{n\Theta_\mu} \;\leq\;E_{n\Theta_\rho} \quad\mbox{for any}\quad
  \Theta_\rho,
\eeq
since
$\displaystyle
  e_{k\Theta_\mu}(x_{<k}) \!=\!
  1\!-\!\mu(x_{<k}\pb x_k^{\Theta_\mu}) \!=\!
  \min_{x_k}(1\!-\!\mu(x_{<k}\pb x_k)) \!\leq\!
  1\!-\!\mu(x_{<k}\pb x_k^{\Theta_\rho}) \!=\!
  e_{k\Theta_\rho}(x_{<k})
$
for any $\rho$.

\subsection{Error Bound}
Of special interest is the universal predictor
$\Theta_\xi$. As $\xi$ converges to $\mu$ the prediction of
$\Theta_\xi$ might converge to the prediction of the optimal
$\Theta_\mu$. Hence, $\Theta_\xi$ may not make many more errors
than $\Theta_\mu$ and, hence, any other predictor $\Theta_\rho$.
Note that $x_k^{\Theta_\rho}$ is a discontinuous function of
$\rho$ and $x_k^{\Theta_\xi}\to x_k^{\Theta_\mu}$ can not be
proved from $\xi\to\mu$. Indeed, this problem occurs in related
prediction schemes, where the predictor has to be regularized so
that it is continuous \cite{Feder:92}. Fortunately this is not
necessary here. We prove the following error bound.

\addcontentsline{toc}{paragraph}{Theorem \ref{thErrBnd} (Error
bound)}
\begin{theorem}[Error bound]\label{thErrBnd}
Let there be sequences $x_1x_2...$ over a finite alphabet $\cal A$
drawn with probability $\mu(\pb x_{1:n})$ for the first $n$
symbols. The $\Theta_\rho$-system predicts by definition
$x_n^{\Theta_\rho}\!\in\!\cal A$ from $x_{<n}$, where
$x_n^{\Theta_\rho}$ maximizes $\rho(x_{<n}\pb x_n)$. $\Theta_\xi$
is the universal prediction scheme based on the universal prior
$\xi$. $\Theta_\mu$ is the optimal informed prediction scheme. The
total $\mu$-expected number of prediction errors $E_{n\Theta_\xi}$
and $E_{n\Theta_\mu}$ of $\Theta_\xi$ and $\Theta_\mu$ as defined
in (\ref{rhoerr}) are bounded in the following way
\beqn\label{th2}
  0 \;\leq\; E_{n\Theta_\xi}-E_{n\Theta_\mu} \;\leq\;
  H_n+\sqrt{4E_{n\Theta_\mu}H_n+H_n^2} \;\leq\;
  2H_n+2\sqrt{E_{n\Theta_\mu}H_n}
\eeqn
where $H_n\!\leq\!\ln{1\over w_\mu}$ is the relative entropy
(\ref{entropy}), and $w_\mu$ is the weight (\ref{xidef})
of $\mu$ in $\xi$.
\end{theorem}

First, we observe that the number of errors $E_{\infty\Theta_\xi}$
of the universal $\Theta_\xi$ predictor is finite if the number of
errors $E_{\infty\Theta_\mu}$ of the informed $\Theta_\mu$
predictor is finite. This is especially the case for deterministic
$\mu$, as $E_{n\Theta_\mu}\!\equiv\!0$ in this
case\footnote{Remember that we named a probability distribution
{\em deterministic} if it is 1 for exactly one sequence and 0 for
all others.}, i.e.\ $\Theta_\xi$ makes only a finite number of
errors on deterministic environments. More precisely,
$E_{\infty\Theta_\xi}\!\leq\!2H_\infty\!\leq\!2\ln{1\over w_\mu}$.
A combinatoric argument shows that there are $M$ and $\mu\!\in\!M$
with $E_{\infty\Theta_\xi}\!\geq\!\log_2|M|$. This shows that the
upper bound $E_{\infty\Theta_\xi}\!\leq\!2\ln|M|$ for uniform $w$
must be rather tight. For more complicated probabilistic
environments, where even the ideal informed system makes an
infinite number of errors, the theorem ensures that the error
excess $E_{n\Theta_\xi}-E_{n\Theta_\mu}$ is only of order
$\sqrt{E_{n\Theta_\mu}}$. The excess is quantified in terms of the
information content $H_n$ of $\mu$ (relative to $\xi$), or the
weight $w_\mu$ of $\mu$ in $\xi$. This ensures that the error
densities $E_n/n$ of both systems converge to each other.
Actually, the theorem ensures more, namely that the quotient
converges to 1, and also gives the speed of convergence
$E_{n\Theta_\xi}/E_{n\Theta_\mu}=1+O(E_{n\Theta_\mu}^{-1/2})
\longrightarrow 1$ for $E_{n\Theta_\mu}\to\infty$.

\subsection{Proof of Theorem \ref{thErrBnd}}
The first inequality in Theorem \ref{thErrBnd} has already been
proved (\ref{Emuopt}). The last inequality is a simple triangle
inequality. For the second inequality, let us start more modestly
and try to find constants $A$ and $B$ that satisfy the linear
inequality
\beq\label{Eineq2}
  E_{n\Theta_\xi} \;\leq\; (A+1)E_{n\Theta_\mu} + (B+1)H_n.
\eeq If we could show \beq\label{eineq2}
  e_{k\Theta_\xi}(x_{<k}) \;\leq\;
  (A+1)e_{k\Theta_\mu}(x_{<k}) + (B+1)h_k(x_{<k})
\eeq
for all $k\!\leq\!n$ and all $x_{<k}$, (\ref{Eineq2}) would
follow immediately by summation and the definition of $E_n$ and
$H_n$. With the abbreviations (\ref{xydef}) and the abbreviations
$m=x_k^{\Theta_\mu}$ and $s\!=\!x_k^{\Theta_\xi}$ the various
error functions can then be expressed by
$e_{k\Theta_\xi}=1\!-\!y_s$, $e_{k\Theta_\mu}=1\!-\!y_m$ and
$h_k=\sum_i y_i\ln{y_i\over z_i}$. Inserting this into
(\ref{eineq2}) we get
\beq\label{detineq}
  1\!-\!y_s \;\leq\;
  (A\!+\!1)(1\!-\!y_m) + (B\!+\!1)\sum_{i=1}^N y_i\ln{y_i\over z_i}.
\eeq
By definition of $x_k^{\Theta_\mu}$ and $x_k^{\Theta_\xi}$ we
have $y_m\!\geq\!y_i$ and $z_s\!\geq\!z_i$ for all $i$.
We prove a sequence of inequalities which show that
\beq\label{detineq2}
  (B\!+\!1)\sum_{i=1}^N y_i\ln{y_i\over z_i} + (A\!+\!1)(1\!-\!y_m)-(1\!-\!y_s)
  \;\geq\; ...
\eeq
is positive for suitable $A\!\geq\!0$ and $B\!\geq\!0$, which
proves (\ref{detineq}). For $m\!=\!s$ (\ref{detineq2}) is
obviously positive since the relative entropy is positive
($h_k\!\geq\!0$). So we will assume $m\!\neq\!s$ in the following.
We replace the relative entropy by the sum over squares
(\ref{entroineqn}) and further keep only contributions from
$i\!=\!m$ and $i\!=\!s$.
\beqn
  ... \;\geq\;
  (B\!+\!1)[(y_m\!-\!z_m)^2+(y_s\!-\!z_s)^2] +
  (A\!+\!1)(1\!-\!y_m)-(1\!-\!y_s)
  \;\geq\; ...
\eeqn
By definition of $y$, $z$, $m$ and $s$ we have the
constraints $y_m\!+\!y_s\!\leq\!1$, $z_m\!+\!z_s\!\leq\!1$,
$y_m\!\geq\!y_s\!\geq\!0$ and $z_s\!\geq\!z_m\!\geq\!0$. From the
latter two it is easy to see that the square terms (as a function
of $z_m$ and $z_s$) are minimized by
$z_m\!=\!z_s\!=\!\odt(y_m+y_s)$. Furthermore, we define
$x\!:=\!y_m\!-\!y_s$ and eliminate $y_s$. \beq\label{detineq4}
  ... \;\geq\;
  (B\!+\!1)\odt x^2 + A(1\!-\!y_m)-x
  \;\geq\; ...
\eeq The constraint on $y_m\!+\!y_s\!\leq\!1$ translates into
$y_m\!\leq\!{x+1\over 2}$, hence (\ref{detineq4}) is minimized by
$y_m\!=\!{x+1\over 2}$. \beq\label{detineq5}
  ... \;\geq\;
  \odt[(B\!+\!1)x^2 - (A\!+\!2)x + A]
  \;\geq\; ...
\eeq (\ref{detineq5}) is quadratic in $x$ and minimized by
$x^*\!=\!{A+2\over 2(B+1)}$. Inserting $x^*$ gives \beq
  ... \;\geq\;
  {4AB-A^2-4\over 8(B+1)} \;\geq\; 0 \quad\mbox{for}\quad
   B\geq\odf A+\odA,\quad A>0,\quad (\Rightarrow B\geq 1).
\eeq
Inequality (\ref{Eineq2}) therefore holds for any $A\!>\!0$,
provided we insert $B\!=\!{1\over 4}A+\odA$. Thus we might
minimize the r.h.s.\ of (\ref{Eineq2}) w.r.t.\ $A$ leading to the
upper bound $$ E_{n\Theta_\xi} \;\leq\; E_{n\Theta_\mu} +
         H_n+\sqrt{4E_{n\mu}H_n+H_n^2}
\qquad\mbox{for}\qquad A^2={H_n\over E_{n\Theta_\mu}+ {1\over
4}H_n} $$ which completes the proof of Theorem \ref{thErrBnd}
$\qed$.

\section{Generalizations}\label{secOut}
In the following we discuss several directions in which the
findings of this work may be extended.

\subsection{General Loss Function}\label{ssecGLoss}
A prediction is very often the basis for some decision. The
decision results in an action, which itself leads to some reward
or loss. To stay in the framework of (passive) prediction we have
to assume that the action itself does not influence the
environment. Let
$l^k_{x_ky_k}(x_{<k})\!\in\![l_{min},l_{min}\!+\!l_\Delta]$ be the
received loss when taking action $y_k\!\in\!\cal Y$ and
$x_k\!\in\!\cal A$ is the k$^{th}$ symbol of the sequence. For
instance, if we make a sequence of weather forecasts $\cal
A\!=\!\{$sunny, rainy$\}$ and base our decision, whether to take
an umbrella or wear sunglasses $\cal Y\!=\!\{$umbrella,
sunglasses$\}$ on it, the action of taking the umbrella or wearing
sunglasses does not influence the future weather (ignoring the
butterfly effect). The error assignment of section \ref{secErr}
falls into this class. The action was just a prediction ($\cal
Y\!=\!A$) and a unit loss was assigned to an erroneous prediction
($l_{x_ky_k}\!=\!1$ for $x_k\!\neq\!y_k$) and no loss to a correct
prediction ($l_{x_kx_k}\!=\!0$). In general, a $\Lambda_\rho$
action/prediction scheme
$y_k^{\Lambda_\rho}:=\minarg_{y_k}\sum_{x_k}\rho(x_{<k}\pb
x_k)l_{x_ky_k}$ can be defined that minimizes the $\rho$-expected
loss. $\Lambda_\xi$ is the universal scheme based on the universal
prior $\xi$. $\Lambda_\mu$ is the optimal informed scheme. In
\cite{Hutter:01op} it is proven that the total $\mu$-expected
losses $L_{n\Lambda_\xi}$ and $L_{n\Lambda_\mu}$ of $\Lambda_\xi$
and $\Lambda_\mu$ are bounded in the following way:
$0\!\leq\!L_{n\Lambda_\xi}\!-\!L_{n\Lambda_\mu}\!\leq\! l_\Delta
H_n+\sqrt{4(L_{n\Lambda_\mu}\!-\!n l_{min})l_\Delta
H_n+l_\Delta^2H_n^2}$. The loss bound has a similar form as the
error bound of Theorem \ref{thErrBnd}, but the proof is much more evolved.

\subsection{Games of Chance}
The general loss bound stated in the previous subsection can be
used to estimate the time needed to reach the winning threshold in
a game of chance (defined as a sequence of bets, observations and
rewards). In step $k$ we bet, depending on the history $x_{<k}$, a
certain amount of money $s_k$, take some action $y_k$, observe
outcome $x_k$, and receive reward $r_k$. Our profit, which we want
to maximize, is
$p_k\!=\!r_k\!-\!s_k\!\in\![p_{max}\!-\!p_\Delta,p_{max}]$. The
loss, which we want to minimize, can be identified with the
negative profit, $l_{x_ky_k}\!=\!-p_k$. The $\Lambda_\rho$-system
acts as to maximize the $\rho$-expected profit. Let $\bar
p_{n\Lambda_\rho}$ be the average expected profit of the first $n$
rounds. One can show that the average profit of the $\Lambda_\xi$
system converges to the best possible average profit $\bar
p_{n\Lambda_\mu}$ achieved by the $\Lambda_\mu$ scheme ($\bar
p_{n\Lambda_\xi}\!-\!\bar
p_{n\Lambda_\mu}\!=\!O(n^{-1/2})\rightarrow 0$ for $n\to\infty$).
If there is a profitable scheme at all, then asymptotically the
universal $\Lambda_\xi$ scheme will also become profitable with
the same average profit. In \cite{Hutter:01op} it is further shown
that $({2p_\Delta\over \bar p_{n\Lambda_\mu}})^2\!\cdot\!d_\mu$
is an upper bound for the number of bets $n$ needed to reach the winning
zone. The bound is proportional to the relative entropy of $\mu$
and $\xi$.

\subsection{Infinite Alphabet}
In many cases the basic prediction unit is not a letter, but a
number (for inducing number sequences), or a word (for completing
sentences), or a real number or vector (for physical
measurements). The prediction may either be generalized to a block
by block prediction of symbols or, more suitably, the finite
alphabet $\cal A$ could be generalized to countable (numbers,
words) or continuous (real or vector) alphabet. The theorems
should generalize to countably infinite alphabets by appropriately taking
the limit $|{\cal A}|\!\to\!\infty$ and to continuous alphabets by
a denseness or separability argument.

\subsection{Partial Prediction, Delayed Prediction, Classification}
The $\Lambda_\rho$ schemes may also be used for partial prediction
where, for instance, only every $m^{th}$ symbol is predicted. This
can be arranged by setting the loss $l^k$ to zero when no
prediction is made, e.g.\ if $k$ is not a multiple of $m$.
Classification could be interpreted as partial sequence
prediction, where $x_{(k-1)m+1:km-1}$ is classified as $x_{km}$.
There are better ways for classification by treating
$x_{(k-1)m+1:km-1}$ as pure conditions in $\xi$, as has been done
in \cite{Hutter:00f} in a more general context. Another
possibility is to generalize the prediction schemes and theorems
to delayed sequence prediction, where the true symbol $x_k$ is
given only in cycle $k\!+\!d$. A delayed feedback is common in
many practical problems.

\subsection{More Active Systems}
Prediction means guessing the future, but not influencing it. A
tiny step in the direction to more active systems, described in subsection
\ref{ssecGLoss}, was to allow the
$\Lambda$ system to act and to receive a loss $l_{x_ky_k}$
depending on the action $y_k$ and the outcome $x_k$. The
probability $\mu$ is still independent of the action, and the loss
function $l^k$ has to be known in advance. This ensures that the
greedy strategy is optimal. The loss function may
be generalized to depend not only on the history $x_{<k}$, but
also on the historic actions $y_{<k}$ with $\mu$ still independent
of the action. It would be interesting to know whether the scheme
$\Lambda$ and/or the loss bounds generalize to this case. The full
model of an acting agent influencing the environment has been
developed in \cite{Hutter:00f}, but loss bounds have yet to be
proven.

\subsection{Miscellaneous}
Another direction is to investigate the learning aspect of
universal prediction. Many prediction schemes explicitly learn and
exploit a model of the environment. Learning and exploitation are
melted together in the framework of universal Bayesian prediction.
A separation of these two aspects in the spirit of hypothesis
learning with MDL \cite{Li:00} could lead to new insights.
Finally, the system should be tested on specific induction
problems for specific $M$ with computable $\xi$.

\section{Summary}\label{secConc}
Solomonoff's universal probability measure has been generalized to
arbitrary probability classes and weights. A wise choice of $M$
widens the applicability by reducing the computational burden for
$\xi$. Convergence of $\xi$ to $\mu$ and error bounds have been
proven for arbitrary finite alphabet. They show that the universal
prediction scheme $\Lambda_\xi$ is an excellent substitute for the
best possible (but generally unknown) informed scheme
$\Lambda_\mu$. Extensions and applications, including general loss
functions and bounds, games of chance, infinite alphabet, partial
and delayed prediction, classification, and more active systems,
have been discussed.


\end{document}